**Title:** An interdisciplinary conceptual study of Artificial Intelligence (AI) for helping benefit-risk assessment practices

**Sub-Title:** Towards a comprehensive qualification matrix of AI programs and devices

**List of authors:**

Gauthier Chassang 1,2,3,4

Mogens Thomsen 1,2,3,4

Pierre Rumeau 5,4

Florence Sèdes 6,4

Alejandra Delfin 1,2,4

**Authors' affiliations:**

1: Inserm, UMR1027, Team BIOETHICS, F-31000, Toulouse, France.

2: Université Paul Sabatier Toulouse 3, UMR1027, F-31000, Toulouse, France.

3: Ethics and Biosciences Platform (Genotoul Societal), GIS Genotoul, Toulouse, France.

4: Unesco Chair: Ethics, Science et Society, Working group: Digital and Robotics Ethics. Université Fédérale de Toulouse (UFT), France.

5: Groupement d'Intérêt Public e.santé Occitanie, Toulouse, France.

6: IRIT-CNRS, Toulouse, France.

**Abstract:**

This paper proposes a comprehensive analysis of existing concepts coming from different disciplines tackling the notion of intelligence, namely psychology and engineering, and from disciplines aiming to regulate AI innovations, namely AI ethics and law. The aim is to identify shared notions or discrepancies to consider for qualifying AI systems. Relevant concepts are integrated into a matrix intended to help defining more precisely when and how computing tools (programs or devices) may be qualified as AI while highlighting critical features to serve a specific technical, ethical and legal assessment of challenges in AI development. Some adaptations of existing notions of AI characteristics are proposed. The matrix is a risk-based conceptual model designed to allow an empirical, flexible and scalable qualification of AI technologies in the perspective of benefit-risk assessment practices, technological monitoring and regulatory compliance: it offers a structured reflection tool for stakeholders in AI development that are engaged in responsible research and innovation.

**Text:**

Several definitions of Artificial Intelligence (AI) have been proposed since the 1950s and the famous works of the mathematician Allan Turing [96] who intended to reproduce human intelligence through automated technological processes and attached testing methods (Turing test [39], [79]) and those of computer scientist John McCarthy currently recognised as the father of modern AI. According to Merriam-Webster modern dictionary, AI is a twofold notion covering, first, "a branch of computer science dealing with the simulation of intelligent behaviour in computers", and second, "the capability of a machine to imitate intelligent human behaviour". Recently, European experts proposed the following definition of AI: "systems that display intelligent behaviour by analysing their environment and taking actions – with some degree of autonomy – to achieve specific goals" [48], [62].

While these definitions might evolve they reveal the final objective of AI, namely mimicking human intelligence. The rapid development of AI-based devices reaching professional and consumer markets (together with a hype about AI capacities) obliges governments to develop, in parallel with industrial strategies, adapted ethical and regulatory frameworks intended to ensure the sole development of AI for the service of humankind and fundamental human rights protection, in the respect of ecosystems.

While AI research and innovation rapidly evolve, often driven by the quest of performance shows, recognition gain and market opportunities, AI conceptualisations are evolving in disciplinary silos and are not structured, voluntarily or not, to enlighten policymakers or regulators, nor to serve the public debate. Today, there is no universal understanding of AI, nor a common legal classification of AI programs and devices. Ethical and regulatory debates start to be organised at international, European, national and local levels with initiatives such as the recent Global Partnership on Artificial Intelligence (GPAI) proposed by Canada and France together with OECD in 2019.Other initiatives are the Artificial Intelligence High Level Expert Group (AIHLEG) created under the auspices of the EU Commission in 2017, and the International Ad Hoc Expert Group (AHEG) of the UNESCO [98], in a quest for appropriate standardisation. But national AI strategies and visions for the field vary and make it difficult to agree on a single harmonised international regulatory framework [66]. To date, more than 90 organisations around the world have proposed ethical principles for AI in the past few years and "leaders of the world's biggest economies – in the G7 and G20 organisations – signed off on a set of extremely vague ethical guidelines" [67] in the early 2019 [58]. One of the regrettable lacks of these initiatives is to directly focus on defining ethical principles without achieving essential upstream steps of qualifications of objects to fall within their scope. As a result, while paramount AI ethical principles are quite clearly established, translation of the principles into engineering practices and laws necessitates tools for defining AI programs and broader AI-based devices or systems in a meaningful way. Envisaging applied ethics and sounded regulations in AI necessitates first an appropriate methodology for conceptualising AI programs and devices in a standardised way based on essential characteristics and functions questioning the ethics and lawfulness of AI programs and devices. Due to the complexity of the field, this conceptualisation effort should be interdisciplinary and pragmatic. It should lead to comprehensive notions and practical tools that could be used by engineers, designers, programmers, policy-makers, lawyers, ethicists, end users and other stakeholders in AI for designing and implementing an ethics-by-design approach. Ideally, AI conceptualisation efforts will not only allow a community to talk the same language, but also ease implementation of specific benefit-risk assessments and help monitoring the technology. For example, regulators will need to legally qualify these technologies for attaching existing values, rules, and perform impact assessments of AI technologies. This could then lead to clearing-house practices and eventually to the creation of new legal provisions. Also, the conceptualisation should facilitate honest and understandable communications towards the public. This is an ambitious and difficult task that needs to reach consensus and potentially new conceptual basis. This could be long and seems unfeasible, but it is a necessary step for avoiding the commercialisation of harmful technologies, as well as the adoption of conceptually unsounded or inapplicable regulations.

Through this paper, we aim at initiating the development of interdisciplinary and practical concepts for qualification of AI programs and devices that could be used by a variety of stakeholders looking for a methodological framework; this will ease the identification of essential features to be considered when envisaging benefit-risk analysis and assessment of an AI-based product. Therefore, we provide a comprehensive picture of existing theoretical concepts emerging from three disciplines, namely Psychology, Cognitive Engineering and Ethics and Law, in order to draw an interdisciplinary common concept that can lead us to a proposal of a qualification matrix for AI programs and devices. This conceptual study will be the landmark for future papers in which we would like to trial the matrix with examples of qualifications of existing technologies in the health sector, a sensitive sector where AI-based innovations and related ethical tensions are numerous and human rights are particularly developed and enshrined into practices. In addition, the health sector is stressing the interaction of AI with human beings in the achievement of a designed purpose. It is a good ground for case-studies. Nevertheless, the reasoning and proposals formulated in this paper could easily be applied to other sectors.

**1. Considering renowned works in cognitive psychology for identifying "intelligence" in AI**

While machines are not humans and do not have natural intelligence, AI was likened to human intelligence early on by its developers. Therefore, yet, human intelligence is an essential conceptual benchmark for AI innovation.

Psychology has been studying human intelligence for many years. Numerous theories have been developed around the concept of intelligence, part of them taking advantage of researches and methodologies used in other disciplines such as anthropology, neurology and even genetics. One of these theories has been developed by Cattell and Horn [65] in the 1960s and further completed, notably by Carroll [13], in order to conceptualise human intelligence and related cognitive abilities, factors, processes and tests. Intelligence (G) refers to the human cognitive abilities and necessary knowledge for learning, solving problems and meeting certain objectives. While the domain is complex and relies on the study of conscious and multitask biological entities, mainly in humans, an interesting general theory and definition of two types of intelligence, namely the fluid (Gf) and crystallised intelligence (Gc), provide useful elements in the context of AI conceptualisation. Any human being has both types of intelligence. Both types can be used together or separately, fully or partly, depending on the situation and functional objectives to accomplish and depending on the initial functional capital of the human brain (initial program set up in AI).

**Crystallised intelligence (Gc)** is defined by Cattell as "the ability to use learned knowledge and experience". For example, this type of intelligence allows answering a problem encountered and appropriately solved in the past, by reproducing equivalent action. It includes the scope and depth of acquired knowledge, the ability to reason by using acquired experiences or procedures and the ability to communicate knowledge. Learning abilities are generally supervised and include encoding capacity, middle-long term memory and ability to access memorised data in a logical way. But learning functions are limited. It is possible to search and select relevant information through the reference database for coping with a known situation or to make correlations between different datasets in order to find appropriate solutions, but without possibility to interpret the datasets in a way that would create new solutions, new strategies. Usually, acquisition or capture of reference data is guided. Gc does not include ability to generalise knowledge. In brief, the approach is based on an identified pool of skills, strategies and knowledge representing the level of cognitive development of an individual through his learning history. Gc is also referred to as "verbal intelligence".
Concrete examples of Gc: being able to use semantic database and vocabulary to formulate correct sentences; being able to read learned language; being able to find similarities between different familiar objects.

**Fluid intelligence (Gf)** is defined as "reasoning ability, and the ability to generate, transform, and manipulate different types of novel information in real-time" [109]. This includes "the ability to solve new problems, use logics in new situations, and identify patterns" without having necessarily the prior experience of similar information or problems or prior strategic learning. Fluid intelligence is thus independent from learning, from experiences, culture and education (in AI we would refer to programming instead of education). Fluid intelligence is shaped by primary abilities such as induction, deduction, linkage, classification and depends on factors such as the processing speed, operative memory and, for a biological entity, neuronal network activity. This type of intelligence has often been compared to the hardware for intelligence as it covers aspects supporting the development of future abilities of an individual. Fluid intelligence includes adaptive skills and abstraction abilities. Extensive learning abilities including broad or specific capture, integration (organisation, categorisation), interpretation, comparison, understanding and retranscription of novel external, environmental data in a comprehensive form are present. Gf can function on the basis of limited working memory [4, 28] (attentional system used both for data processing and for storage based on short-term memory), depending on the quality of the learning ability (in AI we would say learning algorithm) and of data retrieval capacity for problem resolution (working memory can appeal to long-term memory data for improving data processing and can be diminished by parasite stimuli creating attentional troubles, informational or functional overloads). Acquired experiences are usually integrating Gc for feeding the reference database. However, in AI, it is possible that a program functions only on the basis of characteristics related to Gf, or Gc concepts. Gf is also referred as "logico-mathematical intelligence". Concrete examples of Gf: ability to understand natural language, to apprehend implicit meaning or ambiguity for designing appropriate action; ability to interpret novel environmental data, from a simple term to a complex phenomenon, and to make deduction as to their meaning for finding solutions.

In sum, Gf is more "instinctive" than Gc that comes from "know-how". If fluid intelligence is compared to the hardware of intelligence, crystallised intelligence would be the software [75]. To our mind, crystallised and fluid intelligence must remain separated. But both types of intelligence are complementary, some tasks will only need Gf abilities, others only Gc ones, and others will only be possible by combining Gf and Gc abilities. From an AI perspective, instinctive traits related to Gf can be approximated to "autonomous" automated processing leading to new data acquisition and new learning based on data initially unknown by the program. Traits related to the know-how are more related to those AI systems dependent on human programming and benchmark modelling. Gf and Gc include both encoding and data processing abilities.

Carroll (1993) developed further classification by identifying 3 intelligence strata [13]. According to many scholars, Carroll's "review of more than 400 data sets provides the best currently existing single source for organising various concepts related to intelligence" [82]. Fig.1 represents his concept, each stratum being cumulative. Stratum 1 represents specific or narrow cognitive abilities, stratum 2 represents derived broad cognitive abilities, and stratum 3 represents general intelligence, known as the "G factor" according to Charles Spearman [17], and defined as "the fluid ability to integrate multiple cognitive abilities in the service of solving a novel problem and thereby accumulating crystallised knowledge that, in turn, facilitates further higher-level reasoning" [106].

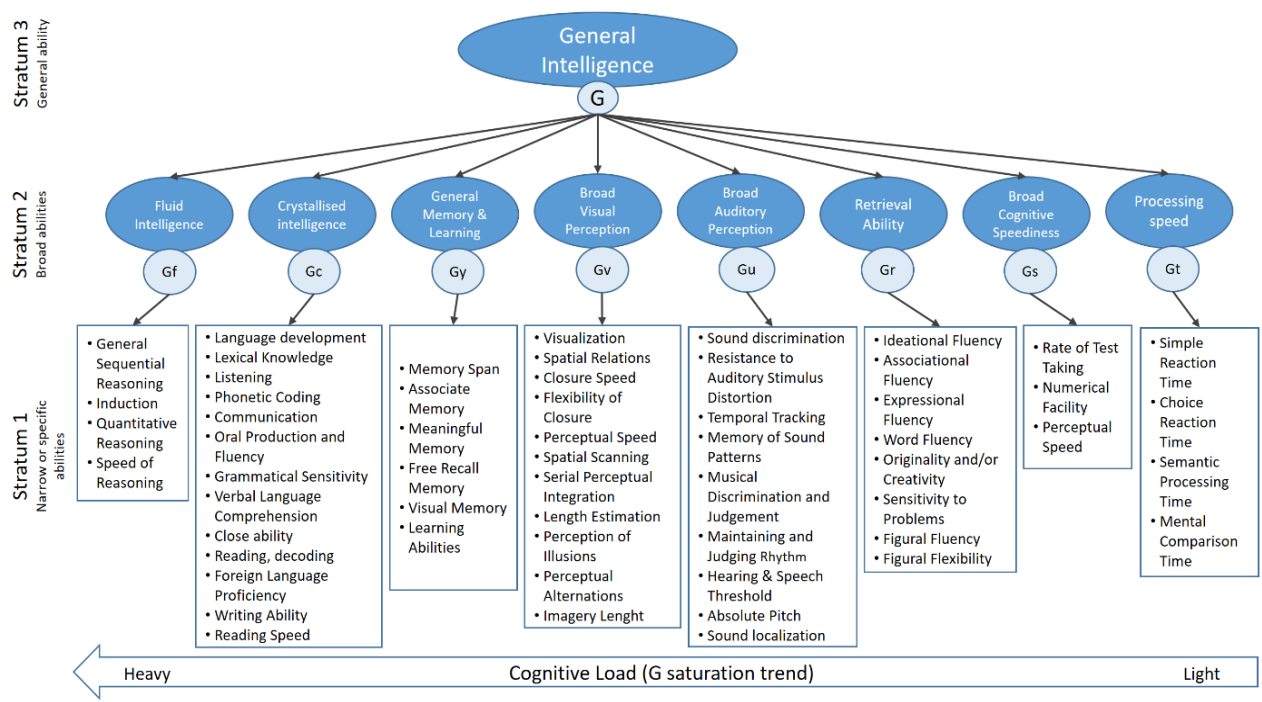

Fig.1: Carroll's Model of Intelligence. Adapted from G.L. Canivez, & E.A.Youngstrom [12].

Nevertheless, in AI, only the distinction between Gf and Gc (Stratum 2) seems crucial and conceptually useful [30]. The other features identified by Carroll in Strata 1 and 2 such as processing speed, visual perception, spatial scanning, sound localisation, are technical features translated into hardware and software algorithms, and are finally accessory and not so determinant for identifying the essence of AI. Similarly, regarding other explored psychological theories, such as the Gardner's theory of "multiple intelligence" [1] [56,16], the study of G factors being an open and evolving research field in psychology [69], we considered only basic concepts applicable to AI. Nevertheless, such aspects are considered through various psychometric tests as intelligence factors. In AI, they could reveal specific risks leading to question proportionality of the accessory functionalities of a program and device, for adapting technological capacities to what is strictly necessary to meet an identified ethical AI purpose.

For simplifying without contradicting the heart of the psychological theories on human intelligence presented above, we will retain the Cattell-Horn's initial distinction between Gc and Gf essentially focusing on learning, memory, search and use abilities. Their approach allows fine-tuning the global understanding of intelligence factors and could inspire the naming of two broad categories of so-called "intelligent agents" composing AI programs, namely "Crystallised" and "Fluid" AI programs, and by extension, linked devices and software. The added value of such an approach is notably related to the representation that such a naming evokes to lay people. Without having performed an extensive study on this, we tested it with people in our direct environment with no specific skills in AI, and without entering into conceptual nor technical details. In short, this naming essentially evoked reflections on the human control of AI programs and devices. Crystallised AI tended to be understood as a guided version of AI, more controllable by humans, more dependent on humans. On the contrary, Fluid AI tended to be understood as a more autonomous AI, based on advanced abilities, *a priori* necessitating less human interventions. Fluid AI shows creativity, in finding solutions to a problem, adaptability to various environments, what evoked more cautions regarding such programs and devices regarding the perceived potential risks of applications. While this would need more research, it seems to be a good starting point for better communicating on AI towards various publics, including lay persons, at least to capture the essence of underlying complex AI systems.

The psychologist Bloom's taxonomy classifying human cognitive processes is also informative to understand intelligent abilities in the meaning of thinking behaviours. [Fig.2 summarises them as they have been lastly completed by the doctrine [52], and analysed in the field of AI [84]. We also adapted it in order to integrate the fundamental concepts of Gc, Gf explored, and to better characterise where a system can be considered mature enough to calculate or compute reasoned decisions. Levels presented in [Fig.2 are cumulative, each building on acquired characteristics of previous levels.

---

[1] Gardner, in the 1980s, distinguished eight types of human intelligence namely visual-spatial intelligence, linguistic-verbal intelligence, intrapersonal intelligence, logical-mathematical intelligence, musical intelligence, bodily-kinesthetic intelligence and naturalistic intelligence. Nevertheless, Gardner's theory has been criticised due to its broadness and lack of empirical studies, but it remains popular and illustrative of human talents, personality traits, and abilities.

| | Lower order thinking skills ───────────────────────▶ Higher order thinking skills | | | | |
|---|---|---|---|---|---|
| **Crystallised Intelligence (Gc)** | | | **Fluid Intelligence (Gf)** | | |
| Level 1 | Level 2 | Level 3 | Level 4 | Level 5 | Level 6 |
| **Remember** | **Understand** | **Apply** | **Analyse** | **Evaluate** | **Create** |
| **Information storage**<br>• Encoding<br>• Knowledge base keeping<br><br>**Recognizing**<br>• Identifying<br><br>**Recalling**<br>• Retrieving | **Interpreting**<br>• Clarifying<br>• Paraphrasing<br>• Representing<br>• Translating<br><br>**Exemplifying**<br>• Illustrating<br>• Instantiating<br><br>**Classifying**<br>• Categorising<br>• Subsuming<br><br>**Summarising**<br>• Abstracting<br>• Generalizing<br><br>**Inferring**<br>• Concluding<br>• Extrapolating<br>• Interpolating<br>• Predicting<br><br>**Comparing**<br>• Contrasting<br>• Mapping<br>• Matching<br><br>**Explaining**<br>• Constructing models | **Executing**<br>• Carrying out<br><br>**Implementing**<br>• Using<br><br>**Decision-making**<br>• Guided response | **Differentiating**<br>• Discriminating<br>• Distinguishing<br>• Focusing<br>• Selecting<br><br>**Organising**<br>• Finding coherence<br>• Integrating<br>• Outlining<br>• Parsing<br>• Structuring<br><br>**Attributing**<br>• Deconstructing<br><br>**Decision-making**<br>• Guided response | **Checking**<br>• Coordinating<br>• Detecting<br>• Monitoring<br>• Testing<br><br>**Critiquing**<br>• Judging<br><br>**Decision-making**<br>• Guided response | **Generating**<br>• Hypothesizing<br><br>**Planning**<br>• Designing<br><br>**Producing**<br>• Constructing<br><br>**Decision-making**<br>• Innovative, unguided response |
| **Evaluation criterion:** Given answer is identical to the one that shall be memorised. | **Evaluation criterion:** Given answer has the same meaning than the information to reformulate. | **Evaluation criteria:** The imposed rule has been respected and the answer is correct (based on a unique answer possibility). | **Evaluation criteria:** Given answer is correct and complete | **Evaluation criteria:** Given answer is relevant, scientifically viable and original | **Evaluation criteria:** Given answer is relevant, scientifically viable and original |

Fig.2: The cognitive process dimension, revised Bloom's taxonomy of learning. Adapted from Anderson and Krathwohl (2001), [1, p. 67-68], and J. Parker and S. Jaeger (2016) [84].

Bloom's taxonomy of cognitive process could be completed by emotional skills (Bloom's Affective Taxonomy) and psychomotor skills (Simpson's Psychomotor Taxonomy) serving the simulation of human intelligence behaviours, notably in robotics. Parker and Jaeger explore these aspects together and concluded that "machines were able to attain about half the learning abilities that humans have" and "none of them possessed all of them" [84, p.11-12]. Interestingly, the authors note that they have "seen that in some areas AI outperforms their human counterparts in the lowest rungs of the taxonomy. However, machines still have a long ladder to climb before reaching the top". Bloom's Taxonomy classification adds another interesting level of specification for achieving a comprehensive qualification of AI.

Psychology researchers also studied the **characteristics of human autonomy**, as a part of human self-determination. This concept deserves great attention due to the regular claims from industrials having developed "autonomous" AI devices and the impact that self-determination of AI devices would have regarding ethics and law if it was proved true. In common language, autonomy is defined as the "ability of someone to be independent, to not be dependent on others; the character of something that works or evolves independently of something else" [73]. Autonomy supposes to be able to act on its own and to appeal to intelligence to find a way to achieve goals or tasks usually self-determined by the individual (by AI in our context). Autonomy is an essential component of self-determination that is a broader concept implying the expression of personality traits through autonomy's features of self-regulation, of empowerment and of self-implementation abilities, thus allowing an individual to notably express his preferences through choices, to adapt and eventually refrain from behaviours contradicting personal values. Therefore, the autonomy concept seems twofold. First, it includes a capacity of something to self-regulate, this eventually leading to action going against the established rule, or against reason (reckless action); second, it includes a capacity of something to react, resist or adapt to an external environment.

While differences exist between psychological and legal theories[2] of autonomy it is interesting to refer to the **three categories of human autonomy** distinguished in psychology [87], namely the "basic autonomy", the "functional autonomy" and the "general autonomy", each being **expressed in two spheres**, in "executive" and "decisional" autonomy spheres, and **through different modes**, namely "direct autonomy", "assisted autonomy" and "delegated autonomy". In AI, the notion of "autonomous devices" is highly controversial and we are not pretending here to solve the issue. But with some adaptation and simplification, by taking AI programs or devices as the subject of analysis, we could consider relevant to distinguish between a machine with "basic autonomy", "functional autonomy" and "general autonomy" to question the ethics of AI devices. Calked on human psychology theory, a machine with **"basic autonomy"** is limited to actions or decisions strictly necessary to maintain its functions. In other words, such a machine could self-maintain or self-repair. A machine with **"functional autonomy"** can perform tasks necessary to community life. In other words, the machine is able to have social relationships with its pairs, between machines, or with humans. Finally, **"general autonomy"** corresponds to other activities necessitating autonomy without being determinant for the autonomous status of the machine. For example, the ability to self-entertain, what a machine cannot reasonably do. Interestingly, from a legal point of view, we could consider adapting relevant distinctions initially made in psychology between **"direct autonomy"** expressed without any intermediary, whether human or material, the highest degree of autonomy yet unachieved in AI, **"assisted autonomy"** expressed with the assistance of an environmental management device intended to increase, amplify, expand, regulate or distribute the effort made by the AI program or device and **"delegated autonomy"** where actions and decisions are totally entrusted to the AI by an external agent, here humans, on a case-by-case or task-by-task basis. Interestingly, questioning the autonomy leads to appraise the **dependence level**. Applied to AI devices, the dependence level should be envisaged first as regard to human intervention on machine functioning and decision-making, then, as regard to the natural/technological environment in which the device operates. Psychologists involved in autism [10] identified three levels of dependence based on the assessment of defined behavioural items characterising human autonomy (known as ESDA[3], [9]) which could inspire AI stakeholders: 1- partial and discontinuous dependence; 2- partial and continuous dependence; 3- full dependence. This scale could be used regarding general functioning of a device or regarding specific characteristics such as cognitive abilities' implementation. It allows notably to identify a system's incapacities or disadvantages in a particular activity and to envisage human interventions, oversight, or further technological tricks for palliating these impediments to autonomous behaviour.

But in psychology, as well as in law, autonomy supposes willingness from the human subject, an essential ability for being free and responsible. Willingness of a machine is not acceptable, the machine functioning on the basis of advanced directives and methods designed by humans and integrated within their program (in software) or command laws (in robots). The machine does not express its own will but reproduces existing and limited schemes in several contexts more or less known in advance. Therefore, an AI system is not really able to govern its own behaviour. But appearances could easily mislead human observers and make them wrongly believe that the machine acts intentionally. Another limit of the concept of autonomy applied to AI devices is the notion of **independent action.** This particularly "refers to the properties of an entity that is capable of operating independently, without being externally controlled or without inputs (material, energy, etc.) from outside" [103]. In practice, any machine, even the more advanced, is dependent on external energy supply, on maintenance operations or on other needs such as an internet connection, a human-based support service. Therefore, so far, no AI-based machine can be considered independent of humans nor autonomous. But innovation goes fast and it is not excluded that real machine autonomy could be achieved one day, at least regarding energetic autonomy if not decisional one. For example, theoretically, a machine could be made independent either through renewable energy sources or perpetual embedded sources, or through dedicated resource-finding, extraction or crafting functions, achieving thus basic autonomy. An important question to address in the perspective of an ethical management of learning machines relates to **machine temporality**, in particular to machine sustainability. Indeed, if the rate of consumption of a finite energy source is so small that the machine would function longer than the average human life we could legitimately argue that such a machine reaches a particular level of autonomy in this regard, triggering special ethical and legal reflections to consider in relationship with other autonomies such as decisional autonomy. Another issue related to AI program or device temporality question their potential to self-improve their autonomy along the time, and to self-direct towards the famous and controversial singularity[4] point where AI systems would be self-sufficient and likely more difficult to control. Kraikivski notes "that machine intelligence can start growing with an initial intelligence capacity that is substantially lower than human intelligence if certain rules for the intelligence capacity growth are implemented in the machine algorithms. These rules include the ability of an algorithm to learn patterns from data, acquire new learning skills and replicate itself. The current computing technologies support the implementation of all of these rules in the form of autonomous algorithms." In such a perspective, even though the author notes that testing could be implemented in order to trial the validity of the singularity theory, he states that "human role should be limited by seeding initial rules for the intelligence growth and providing initial computational resources". The rest should be achieved by the machine. Finally, he concludes stating that "current machine-learning algorithms are already strong in retaining information and recognising patterns in Big Data but weak in deducing new skills and knowledge from it. If artificial intelligence will gain a high capacity in logic, planning and problem-solving, it will still lag behind the human model without the capability of understanding, critical thinking and self-awareness, emotional knowledge, creativity, and consciousness that humans possess. Therefore, building algorithms that are capable of conscious interpretation of perceived information (…) will be crucial for an artificial intelligence to be able to surpass the human intelligence capacity" [72]. Ethical reflections around the appropriateness of such developments should consider the observed decrease in human performances (not skills), based on IQ tests, known as the "negative Flynn effect" theory advanced by Dutton et al. [37] in 2016, even if this effect does not make scientific consensus and is still discussed.

---

[2] Cf. Section 3 regarding the legal notion of "autonomy".
[3] Echelle Stéphanoise de Dépendance dans l'Autisme – ESDA

Nevertheless, it is important to try disentangling complexities around this notion of autonomy in order to serve classification of future AI innovations and advances towards a potential consensus on the (in)existence of "autonomous machine".

This anthropocentric approach to AI based on theories and knowledge of human cognitive psychology could be criticised as it can lead to misleading messages and false expectations about (current) AI [51]. But this is only true if we do not consider the limits of anthropomorphic approaches and the granularity of human intelligence, if we only focus on the General intelligence which, indeed, does not yet exist in a machine. We are aware of these limits and conceptually it makes sense to us to consider these theories as AI pioneers inspired from human brain and neurobiology to create the field, to innovate, by striving to translate and adapt our knowledge on human intelligence to technological contingencies. In such an approach, AI programs can compose the "artificial brain" of a machine. Furthermore, Gc and Gf are not the monopoly of humans and can be found in other living species. And today AI is not only conceived for devices mimicking human behaviour but also animal intelligence or abilities, with the "animats", artificial animals [11]. Therefore, it appears that these core concepts are pretty useful to conceptualise AI basics in terms of abilities related to intelligence even if, taken alone, they are insufficient to shade lights on ethical and legal issues associated with this technology.

---

[4] The singularity refers to an idea that once a machine having an artificial intelligence surpassing the human intelligence capacity is created, it will trigger explosive technological and intelligence growth.

Further elements are necessary to consider in order to envisage tailored qualifications based on a finest typology of AI programs and devices that could lead, then, to legal categories based on legally relevant technical functionalities and effects. For doing so, an exploration of engineering theories is necessary.

## 2. Considering technical features of AI for determining legally relevant functionalities

As seen through Gc and Gf, the learning, the memory and the data capture and processing/reasoning abilities are essential to envisage qualifying an "intelligent" action and to start appreciating the potential of an AI technology. But a specificity in AI is rooted in the human control and intervention in these processes. Cognitive Engineering allows to better appreciate AI specificities in these regards and to question the importance of human action in the AI program and device design and functioning.

In AI there are two big categories of methods used in programming and driving inputs processing through mathematics formula. First, the so-called **Symbolic AI,** which, based on knowledge and logic, processes visual/auditory and other sensor information in the form of symbolic representations of factual events. Symbols, as encoded knowledge, can define things, persons, activities, actions or states, and serve controls of AI applications such as movement, orientation, image recognition and recognition of stimuli, memory and speech functions. The Symbolic AI program is rule-based and causal. Using heuristic approach, it is notably used in expert systems, for medical diagnosis purposes for example, where human knowledge of a problem or of a feature is sufficiently advanced to be precisely described, in our example, medical symptoms. Second, **Statistical AI,** also called "empirical", "connectionist" or "probabilistic" AI, which is based on alphanumerical big data inputs and statistics. Such an AI program is not based on knowledge but on large-scale data from which certain data are extracted and outputs are calculated using statistics. It can serve for example applications performing advanced planning, prevision or recommendation using statistical correlations allowing to calculate or deduce a trend. Symbolic and Statistical AI significantly differ and have both strengths and weaknesses but both are now standard approaches to AI [8]. While the approaches are not exclusive, choosing to rely on one or another can significantly question the robustness of the AI and the quality of its results in a given situation. Top level institutes such as the ANITI in France consider going towards integrative AI, by developing a hybrid AI pattern fed by both approaches for improving cross-understanding of patterns strengths and weaknesses and develop cross-solutions intended to multiply AI potentials and systems robustness [2]. To achieve these methods in AI, European experts mention "several approaches and techniques, such as **machine learning** (of which deep learning and reinforced learning are specific examples), **machine reasoning** (which includes planning, scheduling, knowledge representation and reasoning, search, and optimisation), and **robotics** (which includes, in addition, control, perception, sensors and actuators into cyber-physical systems)" [63, Glossary].

**Machine Learning (ML)**, like its **Deep Learning (DL)** sub-area, is an essential component of an AI program. It has specific technical characteristics basically focused on a simple question: how best to feed the machine's program, as an "artificial brain", to reach "intelligent" action?

In summary, **ML** "is a method where the *target* (goal) is defined and the *steps to reach* that target are learned by the machine itself by *training* (gaining experience)" [18]. The important feature here relates to the training or self-learning function, the latter being understood as the "creation of algorithms which can modify itself without human intervention to produce desired output by feeding itself through structured data" [70]. Four ML characteristics have been identified by scholars, namely **supervised learning, semi-supervised learning, reinforced learning and unsupervised learning** [19]**.** These four methods are based in particular on the necessity of human supervision and on the possibility to use labelled and/or unlabelled data for training the AI program. Most ML methods relies on Natural Language Processing (NLP) algorithms allowing image, text or sound recognition and classification[5]. In NLP the grammatical and vocabulary databases of the program, together with adequate procedures, will determine "the breadth" of the application in terms of data processing. A new, and still problematic [99], category of algorithms emerged as Natural Language Understanding (NLU) [86], going a step beyond NLP by providing post-processing interpretation functions to the AI program (leading potentially to "highly complex endeavours such as the full comprehension of newspaper articles or poetry passages" [104]). In NLU, deep semantic and symbolic skills are crucial, they will represent "the depth" or "the mind" of the application. Some testing of NLU algorithms is already performed in the public internet area, for instance with the last updates of the Google's search engine "BERT" algorithm intended to improve contextual understanding of search queries [80]. **DL** can be defined as "a subset of machine learning where algorithms are created and function similarly to those in machine learning, but there are numerous layers of these algorithms - each providing a different interpretation to the data it feeds on. Such networks of algorithms are called **artificial neural networks**" [70] because their functioning inspires from the function of the human neural networks present in the brain. While there is always a need for human programming, these technical subtleties have consequences regarding the explainability of the algorithmic logic underlying a result, in particular where deep learning is used, this challenging transparency regarding users and some other important legal rules and ethical principles.

---
[5] Watson AI that played TV game Jeopardy.

Therefore, here, it is both the learning method, the use of data that are structured and described by humans (labelled data) and the use of complex artificial neural networks that make the difference in terms of human intervention and explainability of solutions found by an AI program. Also for engineers, as seen earlier, it is foreseeable that advanced ML and DL programs do not need human intervention after initial launching to allow the AI to function. Such AI-based programs and devices could therefore self-instruct, self-behave and eventually self-maintain, provided that they have sufficient decision-making abilities and continuous energy supply.

Based on these learning methods, engineers and scholars highlighted different types of AI-based programs/devices. First, according to their scope, with the notions of **"Weak or Partial"** and **"Strong or Full" AI**. Second, according to their functionalities, with four different **AI Types**.

A starting point for conceiving general categories of AI using such a terminology seems driven by the initial objectives of the AI developers or investors. As Marr schematically notes [77], "generally, people invest in AI development for one of these three objectives: 1. Build systems that think exactly like humans do ("strong AI"); 2. Just get systems to work without figuring out how human reasoning works ("weak AI"); 3. Use human reasoning as a model but not necessarily the end goal" (what we can consider "moderate AI"). This approach based on the intention of AI promoters/developers does not really inform about the technological processes at stake which allow inscribing an item within the AI realm and within a special category. Therefore, we are not retaining the first intention of promoters/developers as a critical criterion for classification, but it is an element to consider and eventually readjust throughout the benefit-risk assessment. This could also play a role in judicial trials, notably against misleading commercial claims.

**Weak / Partial AI** programs are "rooted in the difference between supervised and unsupervised programming. Voice-activated assistance and chess programs often have a programmed response" [71]. In other words, Weak AI depends on programming (benchmarks, targets or objectives, functions) and classification. "Here each and every possible scenario needs to be entered beforehand manually. Each and every weak AI will contribute to the building of strong AI" [18]. It presents a "human-like experience" but is just a simulation as it is usually focused on narrow tasks. Therefore, for example, Siri (Apple) and Alexa (Amazon) bots are classified as Weak AI programs. Likewise regarding the chess program from IBM that beat Garry Kasparov in the 1990s for which all rules and moves were programmed in the machine. Weak AI also includes a notion of narrow AI programs as monotask program. **Strong / Full AI** programs are those tending to fully mimic human brain functioning by using "clustering and association to process data", coming closer to machine reasoning and to the highest G notion, as general intelligence, in human psychology. A strong AI program can go beyond what has been programmed for finding new strategies or solutions to reach an objective. Therefore, once initially programmed and launched by humans, these programs could work with a high degree of autonomy. This category tends to cover semi-supervised or unsupervised programs for which the agent has an extended leeway to calculate and find solutions. Here, the simulation of human brain is questioned as it is foreseen that such AI could be multitask and therefore come closer to human intelligence notions [93].

Nevertheless, as noted by AI experts, "there are still many open ethical, scientific and technological challenges to build the capabilities that would be needed to achieve general AI, such as common sense reasoning, self-awareness, and the ability of the machine to define its own purpose" [62, p.5]. Today, there are no proper existing examples of a strong / full AI program but "some industry leaders are very keen on getting close to build a strong AI which has resulted in rapid progress" [18]. Thus we cannot ignore such a category.

Interestingly, AI programs can be classified based on functionalities or technical features allowing us to better classify the wide variety of AI-based programs/devices. Four types of AI have been distinguished [64]. **Type 1 AI** is a "reactive machine"; **Type 2 AI** is defined by "limited memory"; **Type 3 AI** is defined by social cognition, known as "theory of mind" [85, 54]; and **Type 4 AI** by "self-awareness" or "consciousness", a concept still difficult to apprehend [101, 102] and envisaged through different criteria [89]. These categories are relevant because they inform about potential inherent risks of these technologies, thus potentially paving the way towards the definition of basic technical, organisational and legal standards to fulfil for each type, depending on the purpose and context of use. The following table presents and complements the main technical features of the AI program types identified above, each line being relevant for ethico-legal scrutiny. The criteria presented in the table for each type are not exhaustive.

| AI Program Functionalities | Type 1: Reactive Machine | Type 2: Machine with Limited Memory | Type 3: Machine able to appreciate the mental status of others (Theory of Mind) | Type 4: Self-Conscious Machine |
|---|---|---|---|---|
| Use external data to inform action | ✓ | ✓ | ✓ | ✓ |
| Short-term memory (data retention <1 day) | (✓) | ✓ | ✓ | ✓ |
| Medium-term memory (data retention <45 days) | (✓) | ✓ | ✓ | ✓ |
| Long-term memory (data retention >45 days) | - | (✓) | ✓ | ✓ |
| Use past experience to inform action | - | ✓ | ✓ | ✓ |
| Understand and anticipate humans expectations, emotions, beliefs, thoughts | - | - | ✓ | ✓ |
| Interact socially with humans | - | - | ✓ | ✓ |
| Be self-aware and sentient | - | - | - | ✓ |
| Interact with other technological devices (AI-based or not) | (✓) | (✓) | (✓) | ✓ |

**Legend of symbols used:**

−: Not applicable.
(✓): Possibly applicable.
✓: Applicable.

Table 1: A technical typology of AI program based on machine functionalities

Going a level below, whatever the type of AI program, a number of different specific technical functions and mechanic accessories related, for example, to machine vision, NLP, data interpretation, speech translation from text to speech and/or conversely, data structuration, classification, planning etc. can be enshrined within the device. All will need at least one related "intelligent agent" to be designed and integrated for reaching an intelligent behaviour and a satisfactory result regarding initial program/device goals. While they are all contributing to cognitive computing, such as shown in human psychology with Carroll's theory, they are not critical for classification as they are only means to collect data, as inputs, therefore implying by default privacy and data quality issues. Nevertheless, their appropriateness and necessity regarding the purpose of the program/device shall be considered in order to appreciate the proportionality of a device functionality and avoid undue invasiveness[6].

Experts in AI disagree on what should be called an "intelligent agent" or "intelligent machine". Different definitions [88] have been designed by scholars, with more or less broad scopes, all being essentially based on AI technical features. Taking the human intelligence as a benchmark without some adaptations is not necessarily facilitating the answer due to unknown mechanisms involved in intelligent behaviour as an "agent's ability to achieve goals in a wide range of environments" [74]. And AI can contribute to improve understanding mechanisms of human intelligence. Five classes of intelligent agents can be considered, all being composed of sub-agents performing lower level functions and all being potentially combined in a "multi-agent system" for contributing to solve difficult tasks with behaviours and responses displaying a form of intelligence. These five classes are based on perception and actuation capacities [105, 108]. **Simple reflex agents** are based on current perception of the environment ("how the world really is now") and on a "if-then" rule. Because they are functioning on a condition-action rule such agents are quite limited because they are unable to act without the good condition being perceived. **Model-based reflex agents** are based on knowledge-model (data representing how the real world functions) and on a "find the rule" principle. Current perception of the environment is used to find a rule that matches with the observed environment and trigger the corresponding action. These two first forms of intelligent agents can handle partially observable environments. Options for actions are programmed upstream. **Goal-based agents,** are focused on reaching a predefined goal set or a solution that is the closer to the goal or desired state. There, the program has more flexibility to assess the situation, to assess the effect of an action and its distance from the initial goal of the device. The program's decision-making ability will need to choose the most suitable solution among the options available (e.g. GPS systems) to reach the goal. **Utility-based agents** are similar to goal-based agents but includes an "extra component of utility measurements" which makes the system able to measure the success rate of an action as regard to the goal. The program can identify the best way to achieve the goal, what is useful where there are multiple alternatives. The agent will have to choose based on a utility map serving to calculate the best way to act. These two forms of intelligent agents require fully observable or at least calculable environments for functioning properly. Engineers face challenges related to uncertainties and refer to the fuzzy logic mechanisms, modelling and programming [92, 90] and to Partial Observed Markov Decision Process (POMDP) [36] allowing a better mapping of complex and dynamic environments for reaching greater granularity in automated informed decision-making. **Learning agents** can use past experience to learn or has a learning component serving to translate observable environment in actable variables, for example, to consider user feedback for analysing and improving system's performance. Such agents can be integrated as companion agent of other classes of agents. Depending on the purposes of the AI program and final device, agents can be more or less complex and appeal to abilities and factors associated with crystallised or fluid intelligence to meet their aims. Also, except for simple reflex agents which operate in controlled environment, the other agents are questioning the computer's ability to decide about appropriateness of an action or inaction. This could be driven by logic, by statistics but also by ethical values considered by the human programmer for ethical decision-making.

Any computing device, whether AI-based or not, will take **inputs (data), process** them, and calculate results as **outputs**. There are different issues according to the level of linkage between an AI program and the device actuation in terms of processes used to provide the expected service. The study of the causal link between the AI program and the device output will be particularly interesting from a legal point of view. Technically, challenges for providing or allowing the collection of significant inputs for the system are important, notably for avoiding statistical bias[7]. Similarly, important challenges exist regarding the building of efficient inputs' processing methods and outputs' formulation included in the AI program. Risks of mistakes and injustices through AI use are even greater where the AI device has an important actuator in terms of direct and indirect impacts on the environment in which it operates, this including impact on human rights, calling stakeholders in engineering to pay attention to the guarantees brought through the technique to ensure AI program robustness. Outputs' variety shall also be designed in consideration of their effects regarding environment and in the perspective of always providing human control over the program or device. Testing the device throughout its development and post-marketing to ensure the sustainability of the system's performance, the quality of the results, and for assessing its predictability should help to meet a satisfactory level of control and confidence. This is of particular importance where data inputs are submitted to environmental evolutions and where the device is expected to act autonomously. On this basis, testing procedures and monitoring strategies for preventing functional mistakes or obsolescence of existing safeguards should be further developed for keeping a high quality of the AI system.

---

[6] Cf. Section 3 regarding related ethical and legal challenges.

By considering these elements, it is possible to link psychological theories on Gc and Gf characteristics to technological classifications of AI programs based on learning methods, by inserting further criteria to assess the human role and machine functional autonomy regarding cognitive reasoning and technological abilities to interact with the environment, to self-maintain and eventually to self-represent and self-determine[8]. Keeping this in mind, it is now necessary to address the main legal and ethical issues related to AI programs and devices in order to envisage a precise qualification system allowing to identify relevant regulatory benchmarks for ensuring the protection of human interests and values.

**3. Considering AI legal status for determining adapted legal qualifications and clarifying attached responsibilities**

An essential complement to the previous qualification approach of AI tools is considerations of ethical issues and applicable laws representing the limits of the AI-based functionalities and the necessary options to plan for users in order to respect their rights. Such tools having multiple forms and functions should all be considered from an ethical and legal point of view at several levels, from the program set up to the final legal qualification of the device, in a context-specific approach. As mentioned by the EU Commission and EU AIHLEG, "AI-based systems can be purely software-based, acting in the virtual world (e.g. voice assistants, image analysis software, search engines, speech and face recognition systems) or AI can be embedded in hardware devices (e.g. advanced robots, autonomous cars, drones or Internet of Things applications)" [48, 62]. Let's go through these layers and underline some important elements to consider, in particular within the EU legal and ethical context, with some examples in the health field.

**An AI program,** whatever its characteristics and functions, should be legally qualified as "computer program" (or software), at least in the meaning of EU law [42]. Directive 2009/24/EC, recital 10, states that "the function of a computer program is to communicate and work together with other components of a computer system and with users and, for this purpose, a logical and, where appropriate, physical interconnection and interaction is required to permit all elements of software and hardware to work with other software and hardware and with users in all the ways in which they are intended to function. The parts of the program which provide for such interconnection and interaction between elements of software and hardware are generally known as 'interfaces'." The author of the program and editor of related software can claim copyrights on their creation, as literary works. Conditions attached to such protection claims apply to "the expression of a computer program in any form, but not ideas and principles which underlie a computer or any elements of it" [46]. The program must be original and represent the author's own intellectual creation. The holder of the rights may control the reproduction of the program or of a part thereof, the translation, adaptation, arrangement and any other alteration of the program as well as its distribution. As such, ethical and legal responsibilities related to AI programs development and production rely on the manufacturer's research and development teams including engineers, developers, programmers and designers [76]. It should be performed with due consideration of other disciplines, notably those specifically concerned by the AI program, such as future users' representatives, and with other professionals such as lawyers. AI ethics is a recent field elaborating theories, principles and rules for the AI sector. Considering the ethics of AI algorithms is of paramount importance for reducing risks of misuses or side-effects of the technology before being placed on the market. Performing this ethical assessment is mainly the responsibility of the engineers, developers and designers. The end-users and general public should also be consulted as far as possible and their opinion should be duly considered. Several statements about AI ethics have been issued by different non-governmental and governmental public entities (e.g. Declaration of Montreal [99]; EU [63]) or private organisations (e.g. IEEE [68], ACM [3]). In the EU, the AIHLEG guidelines fix the ethical principles to scrutinise and provide an assessment tool to the stakeholders' community. The high level expert group identifies a global "human-centric" approach to AI developments and seven key requirements that should be met for achieving a "trustworthy AI". "Human-centric" approach "strives to ensure that human values are central to the way in which AI systems are developed, deployed, used and monitored, by ensuring respect for fundamental rights [...]" this including "consideration of the natural environment and of other living beings that are part of the human ecosystem, as well as a sustainable approach enabling the flourishing of future generations to come" [63, p. 37]. The Human is conceived as a referee deciding whether AI action is ethically sounded (desirable), lawful and finally reliable in a specific context (evidence-based approach) this leading to reinforce trust within AI. Therefore, **"trustworthy AI"** is defined by three components. First, it should be **lawful**, ensuring compliance with all applicable laws and regulations. Second it should be **ethical**, demonstrating respect for, and ensure adherence to, ethical principles and values. Third, it should be **robust**, both from a technical and social perspective, since, even with good intentions, AI uses can cause unintentional harm. AI ethics includes horizontal (cross-sectorial requirements) but also vertical aspects (sectorial requirements) to be considered through the design of the program and related device. Several important technical challenges to reach trustworthy AI with embedded ethical algorithms exist [95, 27]. Nevertheless, if human values could not be coded directly within the algorithms, the continued or at least regular human monitoring of AI inputs, processes and outputs ethics must be envisaged as first paramount blocks of safeguards pursuing trustworthy AI.

---

[7] In particular, the bias of selection and endogeneity. Other kind of bias exist and deserve considerations, cf. Section 3.
[8] Cf. Section 4.

A common challenge to the trustworthy AI achievement relies on the intrinsic relationships between humans and AI. Three main categories of biases affect AI: cognitive bias from the programmer, statistical bias, and economical bias [94]. Because of human programming, at least for the initial algorithm, the same **bias** as for human thinking tends to be conveyed through AI programs which will automate them. As BBC mentions, "in short, AI is much more human than we ever realised. Which is perhaps the scariest notion of all" [6]. As an illustration, we can remember the bad outputs from the algorithms of Google [24] and Twitter [57] which showed to be biased and, at use, sexist and racist. Human stereotypes, empirical inequalities and other cognitive bias can easily be translated within the digital AI environment, through the training data used by AI and specific algorithmic instructions of data processing, particularly where the AI program is used in an open environment such as the internet or is used for politics, justice, police and other investigation areas focused on humans, health included [83]. Hence the need to develop upstream human intelligence and understanding of ethico-legal challenges in AI through education, what should ease responsible and ethical AI developments and ethical downstream human checks of AI processing functioning and results, for avoiding automating human bias at large scale [49, p. 11]. In this perspective, the AIHLEG suggests to use "red teaming" to regularly challenge and finally improve the AI programs during Algorithmic Impact Assessments to be implemented throughout the AI system lifecycle [63, p. 20, 37] [see also 26]. The paramount role of ethics in the responsible development of AI programs and devices must be taken seriously. The results of these assessments about the acceptability of a specific AI program could be surprising, in particular if it dramatically improves the respect of human rights while replacing humans. At a certain point, we could also envisage AI to self-maintain ethical behaviour and maybe find new ethical ways of resolving persistent issues. Depending on the ethics referential mobilised, European AI could differ from US, Japanese or Chinese AI. In the field of health and biomedical research, many internationally recognised ethical standards could be used as the common ethico-legal benchmarks for AI developments embracing a global ambition (e.g. UNESCO, WHO, CIOMS, WMA declarations on ethical principles, Council of Europe International Convention and Additional Protocols on Human Rights and Biomedicine and Recommendations). Of note, the UN recently engaged a much needed action and is taking steps with regard to the ethics and regulation of personal health data uses. A first set of Guidelines [97] has recently been published, with a high potential for streamlining AI innovations and protecting international end users in this essential and sensitive field.

**An AI-based system** is composed of three main capabilities according to European experts: **perception, reasoning/decision making, and actuation** [62]. Perception as data acquisition or capture ability can be achieved through various interfaces (e.g. cameras, microphones, keyboards) or sensors (e.g. measuring temperature, pressure, distance, force/torque, rates of certain products, presence or tactile sensors). Reasoning and decision-making relates to the machine's ability[9] to interpret the data in a consistent way in order to envisage an action. It supposes an information processing module intended to make the data understandable for the machine. In any case, AI devices must be legally qualified as data processing[10] devices. Hence existing data protection laws applying to personal and eventually public data [44] must be respected. Actuation is the machine's ability[11] to execute the action it identified as being appropriate to reach a defined goal. Actuation can be performed through actuators, whether physical (e.g. articulated arms) or digital (e.g. activation or denial of a software function) and change the environment, including by producing legal effects regarding third persons or goods. The notion of "robot" [14] is at the heart of reflections about the AI devices. According to CERNA, the robot is a material entity including software. This excludes purely software-based AI such as conversational bots. The capacities of the robot to move, interact and decide by itself are conferring to it a certain degree of autonomy[12]. But as we will see below, such autonomy should be relativized and nuanced when it comes to the legal autonomy notion.

Qualifying the AI-based device in law questions the **legal nature** of the AI device. Is the machine a good (Res) or a person (Personae)? In 2017, the EU Parliament [35] proposed the creation of a digital legal personhood for "autonomous robots", a notion of "an electronic person" consisting in a legal regime close to the corporate personhood applied to legal persons, "where robots make autonomous decisions or otherwise interact with third parties independently". This was including a proposal for a new mandatory insurance scheme for companies producing such devices in order to cover damages caused by the robots, but this controversial proposal has been criticised [34, 78]. Some national advisory bodies close to regulators even departed from the EU report's recommendations (e.g. in France with the OPECST [33]), mainly because such a regulatory perspective would diminish the responsibility of the different persons acting upon the commercialisation of the program or device (programmers, manufacturers, operators). A majority of legal experts definitely tends to consider AI-based devices, including the so-called "autonomous" robots, as legal goods and only goods. Why? Because the robot's full autonomy is a fiction. Daups, based on the difference between technical and philosophical autonomy notions [54], states that, **in law, automated behaviour must not be confounded with autonomous behaviour** [31, 27], see also [32, p. 84-87]. Automated behaviour is rooted in a panel of programmed operations allowing a machine, in defined and limited cases, to execute an action without the need of human intervention. But the action itself and the method is mastered by humans and does not appeal to creativity but to responsiveness. Adaptive ability in responses to environmental evolution is usually not present or not so important due to the controlled environment in which automates are placed. Autonomous behaviour appeals to intelligence, reasoning and enhanced adaptability of the machine's actions outside the normative human control. Based on Daups' proposal in the context of robotic law, a machine's autonomy could be defined in substance as its capacity, in the meaning of the immaterial and material possibility[13], from its own will[14], to decide, in the frame of its competence, about a legal act that is enforceable to thirds and punishable, and whose responsibility can be related to a legal or natural person. The concept of machine autonomy relies on the hypothetic ability of a machine to be able to act appropriately in unplanned situations or unprogrammed scenarios, this being only, to date, a human prerogative. Indeed, machines cannot be considered as having a philosophical autonomy because they cannot create their own law, their own values, or algorithms, nor their own purposes. In fact, even if after initial programming the machine is able to develop its own behavior based on new data acquisition, this will only be based on data recognition, not on plain understanding, not on self- acquired knowledge, it is not entailing willingness nor capacity to self-decide about new values, as matrix rules to perform unprogrammed choices, and this is not constitutive of legal personality or capacity. Without a highest degree of intelligence based on related intelligent agents determining wider potentials for machine's inputs and processing self-design, a machine cannot be presented as reaching full autonomy, as it could not achieve philosophical emancipation regarding pre-existing states allowing to reach a plain self-governing state, a sovereignty. They could be oriented towards such (very hazardous) behaviour by humans, but so far none reached such a philosophical autonomy, remaining thus in a variable level of automating. According to this approach, a machine cannot (yet?) be legally autonomous, implying the impossibility of a machine to perform legal acts[15] by itself in the absence of intrinsic philosophical autonomy and willingness. For instance, a machine cannot provide valid consent on its own behalf. But an AI device can help humans to convey a human-sourced legal act, to formalise or to implement legal acts by being driven by programmed procedures and criteria, in an automated way (e.g. with smart contracts [10]). Therefore, considering a machine as personally accountable is a fiction, the machines being anyway programmed by a human they should only be tools under the human responsibility whose calculated solutions, even if self-determined within a panel of options, are used as a support of human purposes, in human decision-making or into human-controlled environments, under humans' laws such as contract law. While algorithms and AI programs leading to the adoption of a particular solution or pressing humans to adopt a behaviour have a normative effect [5], this effect should not be solely attributed to the machine. It appeals to responsibilities of humans that intervened in the process that led to adopt the problematic solution, this ranging from the manufacturer and programmers to the end-user by passing through persons in charge of the program or device maintenance. Therefore, the machine should not be considered as having a legal capacity, nor as being a person with legal personhood allowing it to be source of legal acts, even if this would simplify the legal reasoning in case of prejudice. Keeping this in mind, because an automated machine can execute legal acts commanded by humans, on their behalf, they can generate involuntary legal facts[16]. These could be related to defects in usual technological processes or be caused by external unplanned or unforeseeable events. In this regard the EU law on general product safety [41] and on defective products [40] establishes horizontal responsibility regimes of interest which could need some adaptation regarding AI, but which already plan relevant legal rules protecting end users. Theoretically, we can highlight a slightly different conceptual understanding of the notion of autonomy in law and in psychology. In law, the focus is made on willingness' autonomy and does not consider intelligence per se. This legal approach would better fit to the broader notion of self-determination defined in psychology, which includes autonomy among other criteria allowing to qualify something autonomous but not self-determined. In psychology, autonomy is focused on the cognitive ability to accomplish tasks, this including the notion of functional autonomy in the task achievement, a notion further explored through engineering regarding technological actuators and machine's material autonomy.

---

[9] Cf. Section 1 regarding the concept of "intelligence".
[10] A data processing can be understood by reference to the definition provided by the EU General Data Protection Regulation 2016/679 (GDPR) Article 4(2) which, by extrapolation, can usefully apply to all data. The GDPR defines data processing as "any operation or set of operations which is performed on […] data or on sets of […] data, whether or not by automated means, such as collection, recording, organisation, structuring, storage, adaptation or alteration, retrieval, consultation, use, disclosure by transmission, dissemination or otherwise making available, alignment or combination, restriction, erasure or destruction".
[11] Cf. Section 1 regarding the concept of "intelligence".
[12] Cf. Section 1 regarding the notion of "autonomy" and its granularity.

As a result, we consider that, at least in the EU legal order, AI devices are first of all legal "Res". They are things, as goods in the meaning of "products which can be valued in money and which are capable, as such, of forming the subject of commercial transactions" [21]. Most of existing AI devices are **assisting human actions**, even when they replace a human factor through task automation, they remain cooperation tools. Robotic autonomy, as well as robotic consciousness, does not exist, and, from a legal point of view, could never exist. Therefore, any personal rights given to existing AI seem themselves legally artificial. The current realistic legal approach of AI programs and devices aims at clarifying and, only if needed, at creating the rights and obligations of the humans behind the machine in order to establish a consistent and practical responsibility regime. This latter will probably prove complex to apprehend at first look as it necessitates to approximate several existing and often complex legal regimes. In our opinion, the driving legal principle in AI regulatory approach should be the subsidiarity of any AI program or device to humans, as a legal good, leading thus to shared competences and responsibility regimes solely attached to humans.

---

[13] Referring respectively to the machine ability of perception and of actuation according to Daups.
[14] Meaning from the machine capacities of appreciation and judgment leading it to select and execute a program deemed adapted to the situation.
[15] Defined as voluntary act that generates legal consequences sought by their author (in our context the machine). For example, contracting.
[16] Defined as a voluntary or involuntary event that generates legal consequences that have not been sought by their author. Examples of legal consequences: creation or modification of rights held by legal persons. For example, the birth and death are legal facts.

This approach is not universally accepted. Indeed, certain States accepted to give some rights to "autonomous" robots. The case of Sophia, a humanoid manufactured by Hanson Robotics and the first robot to officially get Saudi nationality and citizenship on 25 October 2017 marked the century [61]. Similarly regarding the case of a Chinese man who married a robot he created himself [59], a practice foreseen to increase by 2050 [15]. But does this mean that robots shall have rights similar to human beings? Again, this probably does not make sense. We could envisage providing certain rights such as robot integrity and dignity calqued on what already exists for animals for example. Indeed, in many jurisdictions, animals are qualified as "tangible property" on which the owner can have property rights and guardianship duties. In EU countries like in France, animals are recognised as sentient living organisms legally qualified as special tangible goods. Thus, animals have an opposable right to dignity protection [53] obliging humans to provide an ethical attention to animals and to respect them through activities involving animals, this including scientific research [43]. Such a special category also protects animals from humans, from bad treatments and violence. If conscious AI devices, humanoids or animats, exist one day, creating such an intermediary legal qualification and related rights, based on animals' protection regime or on a new one like the electronic person[17], could make sense.

If AI devices are goods and only goods, the next issue that is important in most jurisdictions is to define whether they will be mobile or fixed by nature or by destination. In some contexts, this could be evolutive and an AI device (such as prosthesis) could even integrate the body of an individual becoming then part of the legal person.

Qualification according to the **purpose of the device** brings interesting elements to the reflection. Several actors could have critical role in determining an AI program or device purpose. Defining the purpose of a program and of a broader device are challenges as such. In any case, AI is pursuing OUR objectives, making us responsible of the trajectories and tasks defined and delegated to the machines. The manufacturer of the device, as well as the AI-program developers, engineers and programmers, will be the initial actors in the definition of the task and purpose of their product. First, they need to identify the broad economic sector they are addressing with their creation in order then to determine the technical means to involve and the desired functionalities to include in the AI tool. A broad identification of the economic sector is necessary according to the three sectors identified in economy science [22], namely primary (agriculture), secondary (industry) and tertiary (services) economic sectors. By going a level down, the identification of the targeted precise branch of activity will be necessary. The device could be addressed to actors in different fields as finance, business, law, health, insurance, telecommunication, tourism or research etc. By taking health as an example, the device could be used in administration of health services, for healthcare provision, for health surveillance etc. Each wide sector and branch has legal and sometimes ethical benchmarks to consider for building the algorithm and related program, software and device [27]. Second, they need to inscribe their creation in a specific product group [47] to which specific quality rules are attached. In the EU, as in other parts of the world, the manufacturers must ensure that "products placed on the extended Single Market of the EEA are safe. They are responsible for checking that their products meet EU safety, health, and environmental protection requirements. It is the manufacturer's responsibility to carry out the conformity assessment, set up the technical file, issue the EU declaration of conformity, and affix the CE marking to a product. Only then can this product be traded on the EEA market" [47]. Despite the lack of AI-specific technical quality norm that would include AI ethics check, international quality standards have an important role to play in the global governance of AI as notes P. Cihon [20]. Existing ISO norms, although unspecific to AI, are of interest to ensure materials, products, processes and services quality. For example, AI programs used in self-driving cars will be integrated in the broader legal category of vehicles. An AI program serving military purposes will be legally categorised as a military good, it could ultimately be qualified as weapons, surveillance, tracking or detection devices for example. AI programs used in puppets will be related to toy regulations. Connected devices will also have specific requirements to fulfil. Indeed, AI programs can evolve in various environments including more or less risks regarding users, persons concerned and society. Basically, it can either evolve into a **closed-to-the world environment**, without access to the public internet and usually using a limited number of identified and controlled databases, or into an **open-to-the-world environment** where the program is connected to the public internet, and uses uncontrolled databases, including through the use of observational sensors. No matter whether AI is the driver device or an accessory, quality norms and legal rules attached to their development, commercialisation and uses including limits fixed for protecting public interest will apply to the AI device as a unique good. Questioning the purpose of a program leads to identifying the **target users** of the final product. This is also relevant in the quest for applying a relevant ethico-legal framework. AI devices, including software, can be designed for professionals or consumers. Their leeway in the use of the program or device, including regarding functionalities and purpose modification, shall also be addressed by manufacturers in order to define safeguards limiting potential misuses, by default, such as through Application Program Interfaces[18]. This distributive approach in the qualification of AI products has been recently recognised by the EU and further detailed within the 2020 White paper on AI [49]. Nevertheless, some special features would require more research from a legal perspective.

---

[17] As proposed at the EU Parliament level [35, Point 59]: "creating a specific legal status for robots in the long run, so that at least the most sophisticated autonomous robots could be established as having the status of electronic persons responsible for making good any damage they may cause, and possibly applying electronic personality to cases where robots make autonomous decisions or otherwise interact with third parties independently".

The **modalities of use** of the device are to be considered as an important element. As we said, AI technologies are intended to either assist or replace human action. **Assistance technologies** could be subdivided in two groups based on program/device functional behaviour: executive technologies or vector technologies. Executive technologies perform automated actions fully commanded by algorithms strictly defined by humans to achieve specific tasks (e.g. an algorithm delivering a precise quantity of a product like insulin at determined time intervals). Vector technologies are those that are dependent upon and adaptive to human action (usually physical action, e.g. driver assistance systems or surgical robotic arms). These technologies are assisting/corrective technologies prolonging a human action thanks to AI programs. They are activated on command, and limited to the human action they will act on. **Replacement technologies** can be considered as fully automated programs/devices which do not need humans to work and achieve their goal, to decide, entirely or partly, when to start, when to stop an action, when or how to learn and to implement new way of acting. Decision-making ability is therefore central here. The machine's decision considered here is the one that produces legal effects for humans, users or not, regardless of the result. For being considered as a decisional device, the technology shall produce a judgment based on which action or inaction is implemented. In law, as we have seen, the autonomous character of a decision is an important criterion to distinguish between autonomous and automated decision-making and for attributing this latter to a particular actor allowing identification of corresponding responsibilities or, most probably, responsibility chain. Whatever the supposed degree of AI decisional freedom that could be admitted in certain States, AI decisional action should always engage human responsibilities and be conceived as the result of previous external legal decisions taken by humans regarding the AI device elements (sensors, algorithms…), purposes, terms and conditions of use, promotion, and of internal decision-making technical processes enabling the device. The question will be to determine which of these legal acts generated the situation for establishing responsibilities. Then, for Daups, the central question will be to determine case by case who is responsible for the robot autonomy (and we enlarge his thesis here to machines in general). Daups suggests that, "in fine, the guardian of the autonomy would be the effective responsible of the damage provoked to thirds". The guardian of the autonomy is further defined, in substance, as the legal or natural persons that have the material and the immaterial control of the machine autonomy [32, p.115-120]. Several persons could qualify, notably the end-users, the employers, eventually the device's rental service establishment, but also close or distant operators which acted as machine controllers in case of defective functioning, and which could eventually substitute to users, or the manufacturer of the device, the designers, the programmers etc. Several levels of responsibility could then be determined according to the causality link between their acts/inactions, legal duties, ethical prescriptions, and the damage occurrence. The capacity of the human agents to constrain the level of autonomy of the machine will necessarily be considered to answer the question "who had the authority?". Modalities of use also rely on **subsidiarity games which can be evolving** and crucial. One AI technology could be at a certain time considered as executive and at another time as a replacement technology, this mainly depending on device autonomy. This interplay is normally guided by humans, but we could also imagine autonomous systems such as with airplane piloting AI programs. In certain circumstances a human pilot could be the master, in others, AI could take the lead and eventually keep it even in presence of human-sourced counter indications. It is precisely for regulating such delegations of powers in AI interfaces that AI ethics codes would be important, provided that it is feasible to program ethically. Indeed, such tools should ensure that the AI will take and let the control at the right time, for the right reasons, and to establish "which decision-making procedures can and cannot be transferred to AI systems and when human intervention is desirable or mandatory" [50]. This remains an open research area.

---

[18] An application program interface (API) is a set of routines, protocols, and tools for building software applications. It is a specification of possible interactions with a software component.

AI legal effects should be envisaged through **testing, use-cases and specific benefit/risk assessments**. These assessments should be part of the ethics-by-design and cover, among others, the effects of the use of the final device on user behaviour. Which values do these systems effectively and demonstrably serve? Is the AI device misleading the user in the realisation of a legal act and which are the measures to reduce potential misleading effects (e.g. bias; authority effect…)? Is it creating a disruption on essential elements of human being protection (e.g. an AI used for deciding about granting a right to an individual)? Is the involvement of the AI-based device creating a new legal act as such (e.g. AI self-decision making)? Will the end-user be able to modify the program in order to personalise the use experience or to influence the functioning of the AI device? What is/will be the broader societal impact of the program and device use and spreading? How is privacy affected and respected? These examples of questions should be tackled collegially, interdisciplinary and where possible with representatives of end-users. Identified risks should be accompanied with detailed measures intended to avoid, reduce or eliminate the risks. The efficiency of such measures should be assessed upstream and downstream as part of the quality management process. As far as possible, testing methods, data and assessment results should be made public.

For example, in health, the importance of patient-physician relationship is crucial in many instances due to the fact that only humans can really understand the complexity of humans' behaviour, health status, potential influences on individual choices, and can interact with the patient properly, such as for communicating health information in the respect of biomedical ethics. Nevertheless, in specific instances, the human action is not perceived as essential to protect the rights and interests of the patient. This is the case for performing the medical data analyses, often complex and time-consuming. The literature about AI medical devices shows impressive results demonstrating the added value of using AI compared to human experts in certain tasks (e.g. identification of tumour cells). Even though using AI in clinical settings has the potential to decrease prognostic or diagnostic errors, by considering the complexity or scarcity of certain medical conditions, AI uses could also entail risks of mistakes and lead to wrong conclusions (e.g. due to bias in algorithms, lack of quality of reference data, or automated misinterpretations). Therefore, the appropriateness of the use of AI system must be assessed case-by-case. In certain countries, whatever the use case, sectorial assessments methodologies [60] start to be elaborated. Of note, the human action of competent health professionals is considered essential to review, confirm or eventually infirm solutions found by AI and to inscribe it in real life through concrete measures which will benefit the patient. Such a position affirming the necessity of human control, transparency and auditability of AI systems, in particular where they have a meaningful degree of autonomy, appears essential in a European ethical context [37, p. 9-10] [45, Article 22] and for EU Member States, specifically in health, such as in France, where the position of the National Council of the Physicians' Order (CNOM) [23, p. 20-28 and p. 48-52] and the one of the State Council [25, p. 207] on the matter have been echoed in the current debates on the bill of bioethics law.

Based on a clear classification of AI devices, it is important to consider **AI-based systems** which raise specific issues. According to experts, "trustworthy AI" must comprise "the trustworthiness of all processes and actors that are part of the system's life cycle" [63, p. 5, p. 37]. Defining case-by-case the **actors involved** in an AI system will be important to design shared responsibilities regime. The EU Parliament, through the Delvaux Report [35], identifies device "manufacturer", "the operator" (that can be a distributor, exporter, importer), the owner, the user". This list should nevertheless be completed with some more granularity, in particular behind the term "manufacturer" that should include the AI developers, engineers, designers and programmers who can be different entities from the final manufacturer assembling the device. The notified bodies undertaking quality control over the device prior to the marketing and post-marketing should also be considered as important actors. Advertising companies that can sometimes create hype, false expectations or misleading messages for commercial purposes around AI should also be considered from an ethico-legal perspective. On this basis, existing liability regimes and liability distribution rules can provide an adequate framework for responsible AI development, marketing and use.

In clinical settings, the main actors whose responsibility regarding patients could be engaged depending on the nature of the damage and on the causality link to be established between the incriminated AI actuation and the caused damage are the physician in charge of the medical act, the hospital, the AI device producer/manufacturer and the AI programmer. Each of them has different responsibilities at different levels which could potentially be engaged in the court, case-by-case. In research settings, the research participant could try to engage the responsibility of the researcher, of the research promoter or institute in charge of the management of the specific project action that generated the damage, the AI device producer and the AI programmer. Here again, each of them has different responsibilities at different levels. Interestingly, each actor could look for engaging the responsibility of the others depending on the circumstances. The main difficulty for the patient or research participant side will be to identify the situations where AI devices have been used and to demonstrate a causality link between this tool and the damage. Hence the necessity to enhance transparency about the use of such technology during the informed consent process. Further ethico-legal issues are today well identified such as risks of **discriminations**, AI **explainability and robustness**, including in the field of health [7].

More generally, it should be recommended that AI innovation's stakeholders reflect on the broader societal impact of their AI programs, devices, in a systemic or epistemological approach. This should notably cover ethical issues related to the ubiquitous development of AI, to the product's capacity to modify social relationships, personal identities, relationships with the environment and to e related consequences in short and long terms. Engineers should be able to express their views about the ethics and desirability of a given development to which they contribute, and be protected if they alert the public authorities of any doubtful practices. Policy-makers and regulators should continue paying attention, for example, to new human dependencies to technologies, this including the problem of **excessive trust** in AI-based devices resulting from their uses, and what we would call the AI **imperium effect,** for designating the phenomenon of progressive divestment or abandonment of the human capacity to act to the benefit of complex automated AI systems, the related potential risks regarding AI paternalism on human dignity and psychology [107], as well as the new power games underlying the AI market [91]. Such reflections should serve awareness-raising of all AI actors and complete other targeted assessments performed. They could lead to the elaboration of further risk mitigation strategies, initiatives, collegial opinions, or policies at several organisational levels, from a company to the State.

Ethics by design is necessary but will not suffice alone. Legal safeguards should be clarified for ensuring appropriate protection of human rights as an essential part of the AI consumer new deal, in particular through **post-marketing monitoring**. This includes environmental and social responsibility in AI development and use. In this regard, we presently note the scarcity of relevant labels[19] and related audit processes. However, efforts are ongoing in that way.

---

[19] E.g. Algorithm Data Ethics label (ADEL).

# 4. Proposal of an integrated matrix for qualification of AI programs and devices and instructions for use

The following tables summarise the main elements presented above in a synthetic view. Together, they constitute a matrix for qualifying an AI program/device. Each qualification exercise should be documented and described in writing in order to allow deeper reflections and readjustments. The use of the developments provided above can be useful to complement the qualification and address specific issues.

First exercise consists in inscribing the AI program in one of the broad categories of the following Table 2 based on identified AI program's or device's general purpose, needs, or its already known features. A given program or device can be inscribed in both categories and qualification can be segmented regarding the processes implied for one task and not another. Nevertheless, the main processes at stake in normal program or AI functioning should trigger general qualification of the device (e.g. an AI program only uses Gc mechanisms for a quarter of the tasks while more than half of the tasks operates on the basis of Gf. In such a case the qualification to retain is Gf and, thus, Fluid AI).

| Crystallised Artificial Intelligence | Fluid Artificial Intelligence |
|---|---|
| - Learning abilities (encoding)<br>- Memory abilities<br>- Usage of acquired knowledge/experience abilities to find solutions and base actions (retrieval)<br>- Described as « verbal intelligence » of human being<br>- Limits: cannot serve to find solution to new situations without reference materials related to these new events such as patterns, database | - Learning abilities (encoding + full understanding)<br>- Use, transform, generate, different types of novel information in real-time<br>- Abstraction and Adaptative skills<br>- Independent process from acquired knowledge, experience, patterns<br>- Can function whatever the type of memory abilities<br>- Described as « logico-mathematical » intelligence of a human being<br>- A more instinctive intelligence based on « know-how » |

Table 2: Broad qualification of AI Core Elements according to Intelligence Mechanisms adapted from Cattell/Horn/Carroll's theories.

Then, the goal is to go through each column of the following tables. In some columns, qualification will be alternative (item must enter in one or the other class, exclusively). In other, the qualification could be cumulative (the second class covering automatically aspects of the first one). These modalities are specified at the end of each column by a "A" (alternative mode) or "C" (cumulative mode).

Of note, regarding the assessment of autonomy as a functional behaviour characteristic, we have seen that there is no universal concept of autonomy whether in psychology, in engineering or in law. But each discipline has an entire literature providing criteria and clues allowing to assess and document autonomous behaviour at general and specific level. In sum, the assessors should first tackle separately aspects related to cognitive autonomy (main reference: psychology), to functional and material autonomy (main reference: engineering, robotics) and to legal autonomy (main reference: law) for assessing the program or device abilities and envisage more precise qualification. In terms of functional purpose qualification, by taking the example of a chatbox using AI program for human-machine conversation, the relational purpose could qualify as informational only but could also cover aspects qualifying predictive, propositional of even self-decisional aspects, depending on case-by-case analysis.

Going through the table, it can be assumed that the more an AI program or device inscribes in the lower categories of the table, the higher are risks related to their use.

| | General Category | Cognitive Level | Technological Type | Learning Method | Functional Scope | Functional Previsibility | Functional Behaviour | Functional Purpose |
|---|---|---|---|---|---|---|---|---|
| AI Program or Device | Crystalised AI program | L1 | 1: Reactive Machine | Supervised | Narrow / Monotask | High | Executive | Implemental |
| | | L2 | | | | | | Informational |
| | | L3 | 2: Limited Memory | Semi-supervised | | Medium | Vectorial | Predictive (estimation) |
| | Fluid AI program | L4 | 3: Theory of Mind | Reinforced | Broad / Multitask | | | Propositional |
| | | L5 | 4: Self-consciouness | Unsupervised | | Weak | Autonomous | Self-Decisional (towards users or thirds) |
| | | L6 | | | | | | Conscious acts |
| Modality | A | C | A | A | A | A | A | A or C |

Table 3: Qualification regarding AI product's performances and technical features

The next step is to envisage the ethics and legal aspects related to the product. Each column must be addressed one by one and necessitates specific assessments and developments from the producers, before commercialisation. For a quick explanatory quotation of qualification only the central column of this table can be used. The left and the right columns are indicators of ethical benchmarks that should be adapted to the context in which the AI is developed or to the rules applying in the territory where the program or device will be commercialised.

| | Ethical purpose? | Legal qualification | | | Specific Protective measures |
|---|---|---|---|---|---|
| | EU: « Ethics by design » | 1. Technology | | 2. Sectorial regulations / quality norms | EU: « Human-centric approach » for « Trustworthy » AI |
| Qualified AI program or device | **Beneficence, Sustainability:** « Do good » *(e.g. UN Sustainable Development Goals)* **Non Maleficence:** « Do not harm » **Autonomy of Human being:** « Preserve Human Agency » **Justice, equity, solidarity:** « Be fair and accountable » **Explicability:** « Operate transparently » **Contextual risk assessment** (individual and population levels) | Good (Res) (private/public; related property and responsibility regimes) | Software | Depends on the targetted sectors/users, claims from manufacturer **AND** on General class of the product Technical rules Responsibility rules Examples: - Commercial good for consumers - Commercial goods for professionals *e.g. Medical devices, Personal data processing device, eHealth device, Manufacturing machine, Measuring device, Surveillance or police device* - Military technology - Dual use technology? | 1. Accountability 2. Data Governance 3. Design for all 4. Human oversight 5. Non-discrimination 6. Respect (& enhancement) of Human Autonomy 7. Respect for privacy 8. Robustness 9. Safety 10. Transparency |
| | | Sentient Good (private/public; related property and responsibility regimes) | Hardware (Physical robot such as humanoïds; animats) | | |
| | | Person (Persona) (legal/natural; related rights and freedoms; new digital person?) | | | |
| Modality | C | A | C | A or C | A |

Table 4: Qualification according to identified ethical and regulatory compliance achievements

As a rule, the ethical and legal benchmarks identified and used for the qualification and assessment shall be recorded. Each ethical or legal principle and rule at stake and measures taken to comply with should be documented. Related compliance measures too. At the end of the assessment, stakeholders could decide to qualify their product as ethical and legal.

It is important to remind that the result of the risk assessment performed according to this framework should be available to authorities and made public for ensuring transparency. Also, it should be made clear that in case of important residual risks for users or other persons impacted by the AI program or device which could not be contained or reduced, the stakeholders should refrain from going to the market before finding acceptable solutions. This important rule should be extended to devices which successfully passed the risk assessment, but which reveal defective or disproportionately risky at use. Therefore, we remind here an underlying and unwritten principle of any risk assessment, the **precaution principle**. This latter shall lead to continuous technological refinement for reducing or eliminating risks or to renunciation of the product, this including the product's market withdrawal. This obligation relies first on manufacturers and second on authorities.

Conclusion

**Advantages of the matrix:** The proposed qualification matrix should allow both to reconcile conceptual analyses of intelligence from different disciplines for AI world and allow a deeper analysis of the Crystallised or Fluid AI program or device characteristics, effects on the environment, on human rights, as well as their potentially broader socio-economic impacts, in order to attach relevant ethical and legal frameworks, risk-assessment values, methods and scales. The matrix should help understanding AI characteristics and related technological, ethical and legal challenges. This should serve the design of further risk-assessment tools and risk-control measures of any kind. It could also help identifying gaps in regulations which are necessary to fill in as regard to a new AI-based technology and to eventually create new safeguards for research, commercialisation, and use of such technology. Finally, this study could base further adaptation of foundational concepts of interest for AI in order to (re)tailor basics in the perspective of a consensus building on an intelligible AI classification which would also be useful for communicating with the lay public.

**Limits:** While we did our best to cover and sum up a broad domain through an interdisciplinary approach, we are aware of the limitations related to the completeness of this study and the completeness of the concepts studied. Nevertheless, we strived to provide the roots of a joint approach deserving timely rebirth. Relationships between categories and grey areas exist and further specifications would be desirable through additional research and the creation of new assessment tools. Being also a limit, the compliance with different National specific legislations, on AI as such, if any, or on human and social rights, and cultural diversity of approaches remain big challenges to achieve a consensus on a commonly accepted academic taxonomy for AI.

Acknowledgements

Célia Chassang, Psychologist in child and young adults' development, Specialised in autism, Association InPACTS, France.

Anne-Marie Duguet, Dr in Forensic Medicine, MD.PhD, Emeritus senior Lecturer, Université Toulouse 3 , UMR/Inserm 1027 Unit, Team BIOETHICS, France.

Emmanuelle Rial-Sebbag, Lawyer, Dr in Law, Research Director at Inserm, UMR  1027, Team BIOETHICS, Responsible of the Unesco Chair "Ethics, Science and Society", France.

Catherine Tessier, Researcher and Referent for scientific integrity and research ethics at ONERA - The French Aerospace Lab. Research Director, Expert Engineer, Department DTIS (Traitement de l'Information et Systèmes), France. Member of the UNESCO Ad Hoc Expert Group (AHEG) for the Recommendation on the Ethics of Artificial Intelligence.

This work received the support of the Unesco Chair: Ethics, Science et Society, Working group: Digital and Robotics Ethics. Federal University of Toulouse - Université Fédérale de Toulouse (UFT), France. The authors are responsible for the choice and presentation of the contents of this publication and the opinions expressed therein, which do not necessarily conform to those of UNESCO and do not commit this Organisation.

[106] WISC-V, Second Edition, 2019.

[107] Yonhap News Agency. Interview. Go master Lee says he quits unable to win over AI Go players. Epub, yna.co.kr, 27 November 2019.

[108] Youtube Videos, Lessons on types of AI agents from Education4u and Gate Smashers channels, (Accessed on April 2020)

[109] L. Zaval, Y. Li, E. J. Johnson, E. U. Weber, Chapter 8 - Complementary Contributions of Fluid and Crystallized Intelligence to Decision Making Across the Life Span, in: *Aging and Decision Making*, Academic Press, Editor(s): T.M. Hess, J.N. Strough, C. E. Löckenhoff, ISBN 9780124171480, 2015, p. 149-168.
31